\documentclass{article}
\usepackage{graphicx}
\usepackage{subfig}
\usepackage{booktabs,blindtext}
\usepackage{textcomp}

\usepackage{ijcai17}

\usepackage{times}

\title{How Curiosity can be modeled for a Clickbait Detector\thanks{This work was presented at 1st Workshop on Humanizing AI (HAI) at IJCAI'18 in Stockholm, Sweden.}}

\author{Lasya Venneti\\ 
IIIT, Hyderabad  \\
lasya.venneti@research.iiit.ac.in
\And 
Aniket Alam\\ 
IIIT, Hyderabad  \\
aniket.alam@iiit.ac.in
}

\begin{document}

\maketitle

\begin{abstract}
  The impact of continually evolving digital technologies and the proliferation of communications and content has now been widely acknowledged to be central to understanding our world. What is less acknowledged is that this is based on the successful arousing of curiosity both at the collective and individual levels. Advertisers, communication professionals and news editors are in constant competition to capture attention of the digital population perennially shifty and distracted. This paper, tries to understand how curiosity works in the digital world by attempting the first ever work done on quantifying human curiosity, basing itself on various theories drawn from humanities and social sciences. Curious communication pushes people to spot, read and click the message from their social feed or any other form of online presentation. Our approach focuses on measuring the strength of the stimulus to generate reader curiosity by using unsupervised and supervised machine learning algorithms, but is also informed by philosophical, psychological, neural and cognitive studies on this topic. Manually annotated news headlines – clickbaits -- have been selected for the study, which are known to have drawn huge reader response. A binary classifier was developed based on human curiosity (unlike the work done so far using words and other linguistic features). Our classifier shows an accuracy of 97\%. This work is part of the research in computational humanities on digital politics quantifying the emotions of curiosity and outrage on digital media. 
\end{abstract}

\section{Introduction}

The challenges of quantifying emotions\cite{klaus-et-al:emotion} start with the notorious problem of defining it. The broad consensus definition of emotion encompasses five sub systems of expression, bodily symptoms and arousal, subjective experience (feeling), action tendencies and cognitive, information processing triggers.  The typical approaches of computing emotion are to use range of instruments from pen-and-paper self-rating scales to advanced equipment measuring brain waves and saccadic eye movements, a mix of non-verbal tools, expressive cartoons\cite{Peter:Lixto} which can give objective results free from language \& cultural bias. 

Human curiosity poses similar definitional challenges and has been recognized as one which evokes responses in all the five sub systems, more so it is considered a knowledge emotion, research during 90s concluded that curiosity produces an unpleasant sensation that is reduced by exploratory behaviors. It is believed that rather than serving a purposive end, the primary objective of satisfying one's curiosity is to induce pleasure. The most tangible expression of curiosity is exploratory behavior, by which one tires to satisfy one’s curiosity. 

\subsection{Curiosity mechanisms --- Both aversive and pleasurable}
Humans are deeply curious, need sunlight and knowledge to survive; information seeking has evolutionary roots. We are sometimes called \textquotesingle informavores \textquotesingle \cite{informavores:baits} \cite{Berlyne:curiosity}. The psychological urge evoked by curiosity is accompanied by increased engagement with the world including exploratory behavior, meaning-making, and learning \cite{Kashdan:curiosity}. Berlyne defined two levels of exploratory behaviors, one associated with the perceptual level of curiosity associated with bodily responses, locomotion and investigatory response, and the other associated with the epistemic level of curiosity associated with observation, thinking and consultation. Curiosity is aroused when stimuli contain novelty, uncertainty, complexity, surprise, and conflict between the urges to approach or avoid. Loewenstein proposed the information gap theory for specific epistemic curiosity and its intensity as a function of level of uncertainty generated (follows an inverted U curve).

Spielberger and Starr \cite{Spielberger:Starr} introduced a model of curiosity which can be considered as a positive feeling of interest and wonder (rather than an aversive feeling). Subsequently Litman \cite{Litman:curiosity} incorporated both the motives in to I \& D model calling them I-interest curiosity and D-deprivation curiosity. The curiosity triggered by novelty, surprise or puzzling stimuli viz. diversive perceptual curiosity seems to associated with unpleasant and aversive condition. Here curious exploration is a means to reduce the aversive state of depravation. The other mechanism which embodies our love for knowledge and the drive for its acquisition-epistemic curiosity- is experienced as pleasurable state. Here curiosity provided an intensive motivation of its own sake. Consistent with this perceptual curiosity was found to activate the brain regions that are sensitive to conflict whereas epistemic curiosity activated regions which are linked to anticipation of reward and dopamine neurons responded to information the same way they respond to hunger. The news headline which contains the stimuli to provoke both perceptual and epistemic curiosity in the reader, itself becomes an object of pleasure \cite{Lasya:curiosity}.

\subsection{Related work}
The measurements developed so far are trait curiosity inventories and they explore the desire to gain knowledge or sensory experience.  Perceptual Curiosity Scale \cite{scales:kl-one} was developed to measure the curiosity caused by sensory stimulation (e.g., sights, sounds, odors) Epistemic Curiosity Inventory was developed to measure keenness to acquire knowledge. The two types of curiosity did not show strong correlation indicating two distinct mechanisms. There are other scales developed viz Melbourne Curiosity Inventory  \cite{Naylor:curiosity} for general curiosity, another popular scale was put forth to \cite{Kash:scale} assess curiosity as a function of novelty and absorption and curiosity as a feeling of deprivation scale was proposed. 

The studies done on clickbaits so far are largely based on linguistic features without understanding why such features give rise to reader interest. The contemporary research is based on techniques like forward referencing \cite{Forward-ref:clickbaits}, modeling words \cite{modeling-words:clickbaits}, length \& certain features like pronouns etc. \cite{blendle:clickbaits}, predicting news values \cite{maria:baits} \cite{safran:baits} have also been linked to clickbait headlines.

\cite{potthast:clickbaits} analyzed the features of the clickbaits under 3 heads, viz. Teaser message, linked webpage and meta-information. The classification approach is based on 215 features, which are statistically significant; however, no theoretical explanation was provided for the basis of feature selection and how \& why they work. 

\cite{bhargavi:baits}proposed classification based on sentence structure like length of headline, length of word, syntactic dependency etc. stop words, hyperbolic and common phrases, subjects, determiners and possessives, word n-grams, parts of speech and syntactic n-grams etc though the paper made a passing remark about information-gap and curiosity, however no attempt was made to establish a nexus between the features and role of curiosity in making a headline clickbaity. 

\cite{chen:clickbaits} theorized that curiosity particularly information-gap, is responsible for the clickbaitiness of low quality news, however no computational work has been done except a mention of ‘Unresolved pronouns’ being the curiosity piquing mechanism.

\section{Methodology} 
Measurement of curious trait(of individuals) is not very relevant to stimulate curiosity, the focus is on constructing curious stimuli to arouse curiosity. We therefore attempted to measure the stimulus, a news headline for its strength to get noticed, clicked and read when many other stimuli compete for reader attention.
Therefore, the curiosity value of the headline was computed for two distinct facets viz. Diversive and epistemic as the combination of both result in reader spotting, selecting and clicking the headline. For this task, we have selected trivial headlines called clickbaits (from among other possible domains like sports, TV serials, advertisements, thriller and horror movies, literature etc.) which attracted huge reader attention evidenced by clicking and sharing metrics. While media pundits criticize clickbaits as trickery, continued popularity and growth in the industry warrants a deeper study of reader motivation to click such headlines.

\subsection{Dataset Overview}

The dataset on which the curiosity based classification metrics have been computed has been collated by \cite{bhargavi:baits} consisting of 16,000 clickbaits and 16,000 non-clickbaits. We have taken appropriate lexicons to help in the classification task. We have used the ‘ABC million news headlines’\cite{ABC:baits} dataset, ordered by date having headlines between early-2003 to late-2017. Every event of significance in the given date range has been captured in the dataset, and hence has been used as the reference for the news topics that the general public is aware of in the context of reading news. 

\subsection{Visual Sampling}

The saccadic sampling and oculomotor system is an excellent model system for understanding information-sampling mechanisms and is considered the most efficient way by which humans sample the visual stimuli \cite{saccadic:baits}. Therefore we attempted to measure the collative variables of novelty and surprise which are considered key factors in spotting of headlines.

\subsubsection{Novelty}

To compute novelty of the clickbaits vs non-clickbaits, we specified a Topic Model (Latent Dirichlet Allocation) on the ABC news headlines dataset between the duration of Sep 2014 to Sep 2015, as the headlines from clickbait and non-clickbait dataset fall between this date range. The Topic Model was trained to produce 200 topics, thus covering all the major stories between the given dates. We generated a probability distribution over the 200 topics for each headline in clickbaits and non-clickbaits. We calculated the information distance between the headline topics that the users were exposed to, and the topics that are present in clickbaits and non-clickbaits.  We found that clickbaits are significantly more novel than non-clickbaits using two metrics Hellinger distance(symmetric) and KL-divergence (not symmetric)(Figure1). These distance metrics have been used as the distinguishing features for clickbaits vs non-clickbaits.

Kullback---Leibler divergence is a measure of how one probability distribution diverges from a second, expected probability distribution.\cite{KL:baits}

\begin{equation}
D_{\mathrm {KL} }(P\|Q)=\sum _{i}P(i)\,\log {\frac {P(i)}{Q(i)}}
\end{equation}
Hellinger distance: In probability and statistics, the Hellinger distance is used to quantify the similarity between two probability distributions. \cite{H:baits}
\begin{equation}
H(P,Q)={\frac {1}{\sqrt {2}}}\;{\sqrt {\sum _{i=1}^{k}({\sqrt {p_{i}}}-{\sqrt {q_{i}}})^{2}}},
\end{equation}
\newline
\begin{center}
 \begin{tabular}{c c c c} 
 \hline
      & Mean & Variance  \\ [0.5ex] 
      & C \hfill NC & C \hfill NC \\ 
 \hline
 KL & 2.72 \hfill 2.37 & 0.01 \hfill 0.16  \\
 H & 0.76 \hfill 0.57 & 0.00 \hfill 0.01 \\
 \hline
\end{tabular}
\end{center}

\begin{figure}[h]%
    \centering
    \subfloat[Clickbaits are significantly more novel(metric 1)]{{\includegraphics[width=6cm]{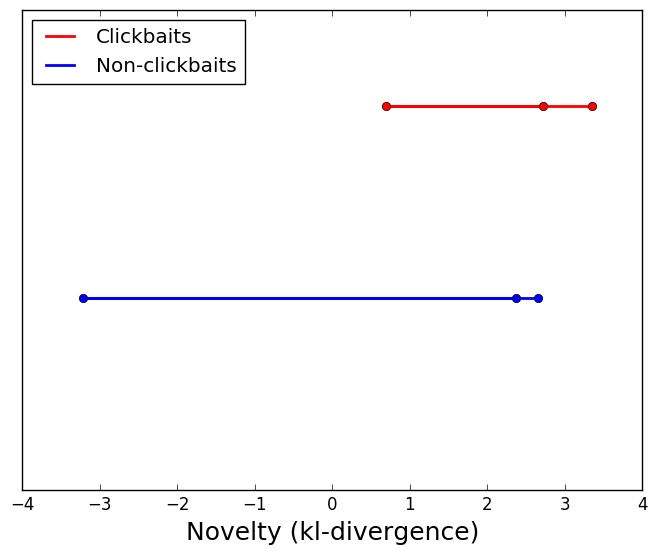} }}%
    \qquad
    \subfloat[Clickbaits are significantly more novel(metric 2)]{{\includegraphics[width=6cm]{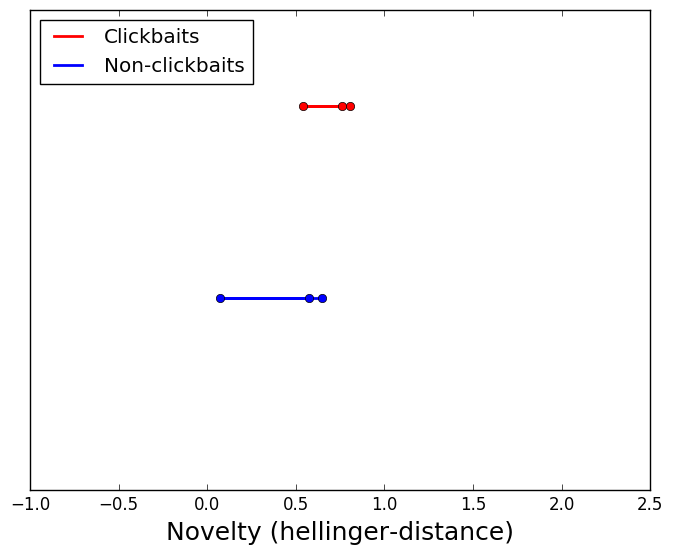} }}%
    \caption{Novelty metrics}%
    \label{}%
\end{figure}
\subsubsection{Surprise}
To compute surprise, we took bigrams of clickbaits and non-clickbaits, and measured the frequency with which they occurred in the ABC news headlines corpus. Thus, each clickbait or non-clickbait can be represented by the frequency of each bigram for the whole sentence, which we call surprise frequency vector. We hypothesize that more frequent the occurence of a bigram, lesser is the perceived surprise value of the bigram when encountered by a reader. We fine-tuned our hypothesis, by formulating that the length of the number of continuous 0s (i.e bigrams that have no occurences in the ABC news headlines corpus) and the greatest non-zero values (bigrams in a clickbait or non-clickbait that have the most occurence in the ABC corpus) in the clickbaits and non-clickbaits surprise frequency vectors, are the distinguishing features of the headlines belonging to the two categories. It is evident that continuous 0s indicate a string of usages which are uncommon, most likely violate the reader expectations thus cause curiosity. Wheras, the greatest non-zero value indicates common usage which is prevalent in non-clickbaits. Using these two contrasting features, we have captured the surprising nature of clickbaits. The figures(Fig 2) show a clear distinction between clickbaits and non-clickbaits, as length of continuous 0s and value of the greatest non-zero number vary.

\begin{figure}[h]%
    \centering
    \subfloat[Non-clickbaits predominantly have smaller length of number of continuous zeros, whereas clickbaits dominate as the length increases]{{\includegraphics[width= 6cm]{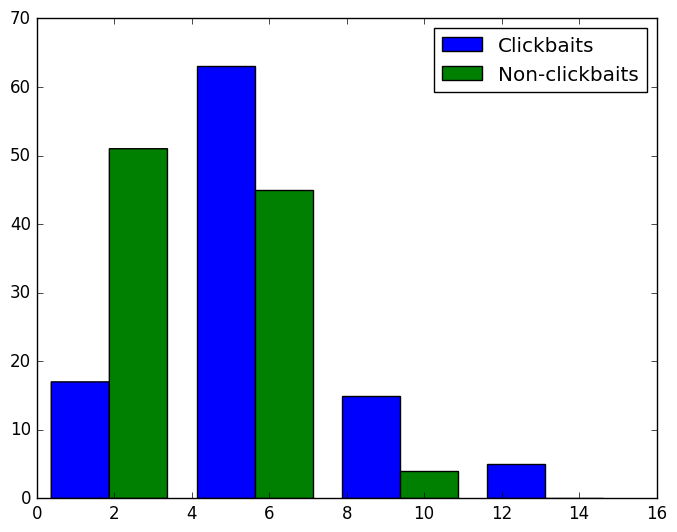} }}%
    \qquad
    \subfloat[Clickbaits dominate the largest values in the smaller range non-zero values in the surprise vectors, whereas non-clickbaits dominate as the value increases]{{\includegraphics[width= 6cm]{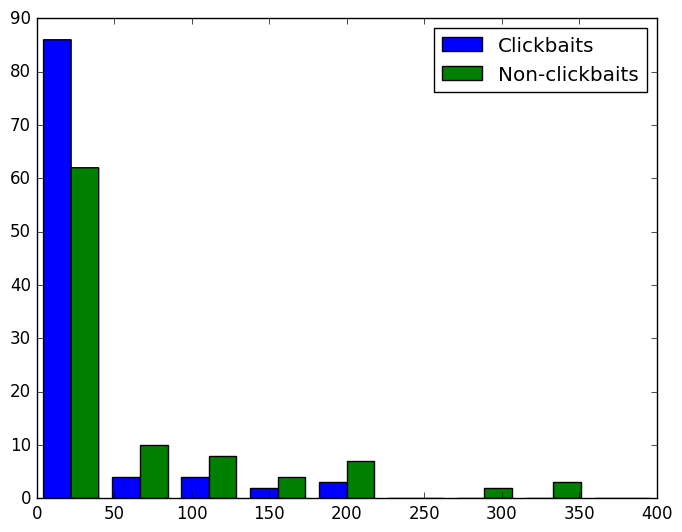} }}%
    \caption{Histogram showing distinguishing features in surprise vectors}%
    \label{}%
\end{figure}

\subsection{Information-gap}

George Loewenstein\textquotesingle s information gap theory \cite{Loewenstein:baits} proposed after the second wave of research, is the most influential one among the psychology approaches. He treats curiosity as reference point phenomenon which gets aroused when one’s informational reference point in a particular domain gets elevated above one’s current level of knowledge. Thus, this approach suggests the possible considerations for curiosity inducing headlines, the situational determinants proposed give us insights about how and why headlines work. We have derived the features of the headlines from the theory and then came out with the computation of the same.

\subsubsection{Reference point elevation}
The information gap opens up when a question gets activated in the mind of the reader. When the headline enables this by using interrogative and question raising constructions and promises a new information the reference point goes up. We used interrogative sense of the sentence and question usages as the distinguishing features. 
We performed semantic role labeling using practnlptools on the headlines, to better understand which agents and actions are leading to reference point elevation, and found distinguishing characteristics, for the labels ARG-TMP(temporal), ARG-MOD(modal), ARG-PNC(purpose). For ARG-TMP and ARG-PNC we found that in clickbaits these arguments less precise and longer, leading to reference point elevation. Whereas, there is higher presence of ARG-MOD in clickbaits, and far fewer in non clickbaits.  

We also employed the usage of \textquotesingle uncertain \textquotesingle words released by \cite{bias:baits} and anticipation words collated as part of the NRC Emotion Lexicon created by Saif Mohammad and Peter Turney at the National Research Council Canada, as uncertainity and anticipation also result in reference point elevation.

The perception of information gap becomes more intense in the reader\textquotesingle s mind when it is perceived to be be important, salient and has explicit/implicit promise of closing the information gap\cite{Golman:baits} .

\subsubsection{Importance of the headline}
This depends on the attention reader gives to the information promised, which is in turn depends on the importance reader gives (circular relationship admittedly). To bring this out we have used the concept of \textquotesingle self \textquotesingle, this is measured by computing the first and second person pronouns, various measures of self and other \textquotesingle self \textquotesingle concepts used are from social psychology viz. reflexive consciousness, interpersonal being and executive function. We identified 14 General Inquirer dictionaries related to the concept. 

\subsubsection{Saliency}
 The information gap caused by a headline becomes salient by the progression of contextual factors. The reader would be more curious with the contextual proximity and this is computed by measuring contextual words, dates, point of time, period of time, continuous and perfect tenses.

\subsubsection{Epiphany}
Readers when made aware of the information gap consider it worthy enough of attention when they perceive comprehensive resolution of various possibilities raised in the question.This is brought out by presence of 'listicles' (the phenomenon of making a list out of a piece of writing) represented by usage of numbers in the headlines. This is widely observed in clickbaits. 
\newline
\section{Results and Discussion}
\begin{flushleft}
 	\begin{tabular}{c c c c c c c} 
 	\hline
 	Model & Features & Accuracy & F1-Score & MSE \\ [0.5ex] 
 	\hline
 	SVM & Novelty & 0.9916   & 0.9915 &  0.033 \\
    SVM & Surprise & 0.9376  & 0.9678 & 0.2492 \\
    SVM & All features & 0.8986   & 0.8978 & 0.405 \\
    \hline
    LogReg & Novelty & 0.9914   & 0.9913 & 0.034\\
    LogReg & Surprise & 0.9369  & 0.9674 & 0.2523 \\
    LogReg & All features & 0.9717 & 0.9713 & 0.113 \\
 	\hline
	\end{tabular}
\end{flushleft}
We have brought out the tool to measure the strength of the headline to generate reader curiosity, using unsupervised and supervised learning techniques,  by using the features described above. We experimented with various methods, and feature sets, and could achieve high accuracy of 97.17\% of classification using all the handcrafted features by taking 1:4 ratio of training-testing data.

Consistent with the curiosity mechanism discussed in section 1.1 and visual sampling discussed in 2.2, the reader (eye) samples the headline for its novelty and surprise value, then piqued by the information gap and activated questions generated in reader’s mind by various stylistic features. These enhance the reader’s need to know little more than the current knowledge levels (thereby making the reader aware of ignorance and marginally elevating the reference point) and various linguistic usages make the information gap self-important, salient with a promise to close the gaps. The computational values brought out that novelty is semantic (topic away from routine, which may have hedonistic overtones). 

The presence of interest value, combined with the stimulus results in arousal of curiosity. To assess whether the topics in click baits, are within the interest of the reader we specified a topic model, and generated the coherence scores for number of topics(N) to be specified to the model. We picked the N value having the highest coherence value.

\begin{center}
 	\begin{tabular}{c c} 
 	\hline
 	Coherence & N  \\ [0.5ex] 
 	\hline
	-17.8481821611 & 62\\
	-17.7726014487 & 63\\
	-17.5700963699 & 64\\
	-17.7370527362 & 65\\
	-17.659071172 & 66\\
	-17.5567553857 & 67\\
	\textbf{-17.480575445} & \textbf{68}\\
	-17.6596784385 & 69\\
	-17.5682784624 & 70\\
	-17.7595639561 & 71\\
	-18.0399170013 & 72\\
 	\hline
	\end{tabular}
\end{center} 

Typical topics of the clickbaits identified by the algorithm show high degree of novelty (we performed qualitative coding on the clusters, the topics could be interpreted as the following: astrology, zodiac signs, Fast Food, beauty \& appearance, socialization, celebrity, gossip, supernatural characters, health \& life style, wellness, music, video, art) compared to news headlines for the chosen period (sports, politics weather, countries, war, negotiation etc.) Clickbait topics are much closer to ideas of self \& identity, popular culture and thus are of interest to the readers. 

Similarly, we found surprise is resulting in a headline from an unconventional or not so frequent usage of word arrangements causing instant gap in the reader’s expectation. 

Information gap is contained more in stylistic features of the sentence clearly indicating that the same topic can be designed to cause curiosity in the reader’s mind. 

Thus headlines cause curiosity, for the novelty value of the topic, surprise by word arrangements and information gap by stylistic features. Our study first of its kind validates some of the core concepts of human curiosity to bring out the reader motivation. The study results show high accuracy and qualitative results are consistent with the published opinions of the media professionals. We could successfully adapt machine learning algorithms to accomplish this, thus furthering scope for AI based interventions in designing communication to attract attention from the target audience.

\subsection{Conclusion}
Cognitive experiments \cite{laura:baits} reveal that purely
pleasurable stimuli sometimes take second seat to maximize learning, to understand cause \& effect relationship, to decrease prediction error, to discover the structure of the world. This would mean epistemic curiosity scores over diversive curiosity. Appropriate computational analysis can be carried out to establish that news headlines
with strong information gap score over purely novelty based headlines.

The theory also evidences the fact that perceptual curiosity decreases with age and specific epistemic curiosity remains stable into adulthood and even in to old age. Thus headlines for the youth and adult mature population needs to be different. Also mature men continue to learn, whereas youth likes novelty and believe in taking
chances. Further work can be done to train algorithms to aid product design and segmenting the target reader group.

\appendix

\bibliographystyle{named}
\bibliography{ijcai17}

\end{document}